\documentclass{article}

\usepackage[sort,numbers]{natbib}
\usepackage[preprint]{nips_2018}
\usepackage[utf8]{inputenc} 
\usepackage[T1]{fontenc}    
\usepackage[colorlinks=false,urlcolor=blue]{hyperref}       
\usepackage{booktabs}       
\usepackage{amsfonts}       
\usepackage{amsmath}
\usepackage{nicefrac}       
\usepackage{microtype}      
\usepackage{graphicx}	   
\usepackage[footnotesize,center]{subfigure}  
\usepackage{authblk}

\newcommand{\figref}[1]{Figure~\ref{#1}}
\newcommand{\tableref}[1]{Table~\ref{#1}}

\title{A Follow-the-Leader Strategy using Hierarchical Deep Neural Networks with Grouped Convolutions}

\author[1]{Jos\'e Enrique Solomon}
\author[2]{Fran\c cois Charette*}
\affil[1]{Graphcore Inc.,   167 Hamilton Ave,  Palo Alto,  CA 94301}
\affil[2]{Greenfield Labs,  Ford Motor Company,   3251 Hillview Ave,  Palo Alto,  CA 94304}

\begin{document}

\maketitle


\begin{abstract}
The task of following-the-leader is implemented using a hierarchical Deep Neural Network (DNN) end-to-end driving model to match the direction and speed of a target pedestrian. The model uses a classifier DNN to determine if the pedestrian is within the field of view of the camera sensor. If the pedestrian is present, the image stream from the camera is fed to a regression DNN which simultaneously adjusts the autonomous vehicle's steering and throttle to keep cadence with the pedestrian. If the pedestrian is not visible, the vehicle uses a straightforward exploratory search strategy to reacquire the tracking objective. The classifier and regression DNNs incorporate grouped convolutions to boost model performance as well as to significantly reduce parameter count and compute latency. The models are trained on the Intelligence Processing Unit (IPU) to leverage its fine-grain compute capabilities in order to minimize time-to-train. The results indicate very robust tracking behavior on the part of the autonomous vehicle in terms of its steering and throttle profiles, while requiring minimal data collection to produce. The throughput in terms of processing training samples has been boosted by the use of the IPU in conjunction with grouped convolutions by a factor  ${\sim}3.5$ for training of the classifier and a factor of ${\sim}7$ for the regression network. A recording of the vehicle tracking a pedestrian has been produced and is available on the web. This is a preprint of an article published in SN Computer Science.  The final authenticated version is available online at: \href{https://doi.org/10.1007/s42979-021-00572-1}{https://doi.org/https://doi.org/10.1007/s42979-021-00572-1}.

\end{abstract}

\section{Introduction}

Autonomous vehicles (AV) and general autonomous systems (AS)  are being tasked in an ever increasing number of applications and within ever more diverse environments. From deep sea exploration to delivering meals in highly congested urban centers \citep{Delivery, AUVs}, they are performing their duties in a number of highly complex ecosystems. These tasks not only require novel capabilities in terms of physical manipulation of the immediate surroundings, but also new control paradigms that permit for rigorous, repeatable, and flexible behavior on the part of the AS/AV. 

The tasks of following-the-leader and platooning are recurring objectives in a variety of these applications, and though of a seemingly mundane nature,  are relatively complex for a computational framework to perform. Various approaches have been documented in the literature. Earlier efforts focused on creating robust kinematic models that require only minimal information from target objects, such as in Klanc\v{a}r et al \citep{laser}, where a Euler integrable system of equations is fed the relative distance and angular trajectory of the target to force adherence to a tracking path. In Baarth et al \citep{Kinect}, a state-space kinematic model is presented that uses a Kinect\textsuperscript{\tiny{\textregistered}} sensor to determine distance to the target object and adjusts velocity and trajectory to maintain a one meter distance while following. The common theme across much of this body of work is to minimize the complexity of the input signal while maximizing the responsiveness of the tracking vehicle by creating detailed mathematical representations that allow for greatest flexibility in kinematic response. The approach taken here is to leverage the representation capabilities of DNNs and use a raw image feed as input and allow for robust kinematic behavior to be learned by the network implicitly. 

The focus of the current effort is in training an AV to follow a pedestrian. The specific driver here is the concept of a technician working on an assembly line that requires either a heavy tool or large part to be carried from one location of the factory floor to a destination not known a priori.  The approach taken is to build on the authors' prior work \citep{MULTI} and train a hierarchical deep learning framework to handle the task. In the hierarchical framework, two distinct networks work in unison: an entry DNN is a classifier model which determines if the target pedestrian is within the field of view and if so, feeds the camera data to a subservient regression DNN which adjusts the vehicle's steering and throttle to maintain cadence with the pedestrian. If the target pedestrian is not detected, then an exploratory strategy is used to reacquire them. Using two distinct networks to subdivide the task allows for more shallow network topology which lowers the computational load on  the embedded compute platform, reduces the amount of data required for training, and helps build a modular system that permits functionality upgrades.

In terms of the networks developed for this work, in order to reduce both time-to-train and inference latency on the AV platform, grouped convolutions, \citep{AlexNet, ResNeXt}, have been introduced into both network structures.  All models developed for the work were trained on the IPU to leverage its fine-grain compute capabilities. This novel graph processor has demonstrated significant differentiation in terms of speed-up when training grouped convolutions since these generally consist of dispersed node connectivity and so benefit from the IPU's Multiple Instruction, Multiple Data (MIMD) compute platform \citep{CITADEL}.

The outline of the paper is as follows. The hierarchical DNN framework is introduced, discussing the primary functional characteristics of how the scaled AV performs its objective. The two DNNs that are integral to the framework are then presented, followed by a conceptual review of grouped convolutions.  The method by which a training set was generated is discussed, followed by a description of the scaled AV platform. The recipes by which the models were trained are then reviewed.  Quantified results for both the classifier and regression networks are presented,  which is followed by a conclusion section.

\section{Hierarchical Framework}

\begin{figure}[htb]
  \centering
  \includegraphics[scale=0.4]{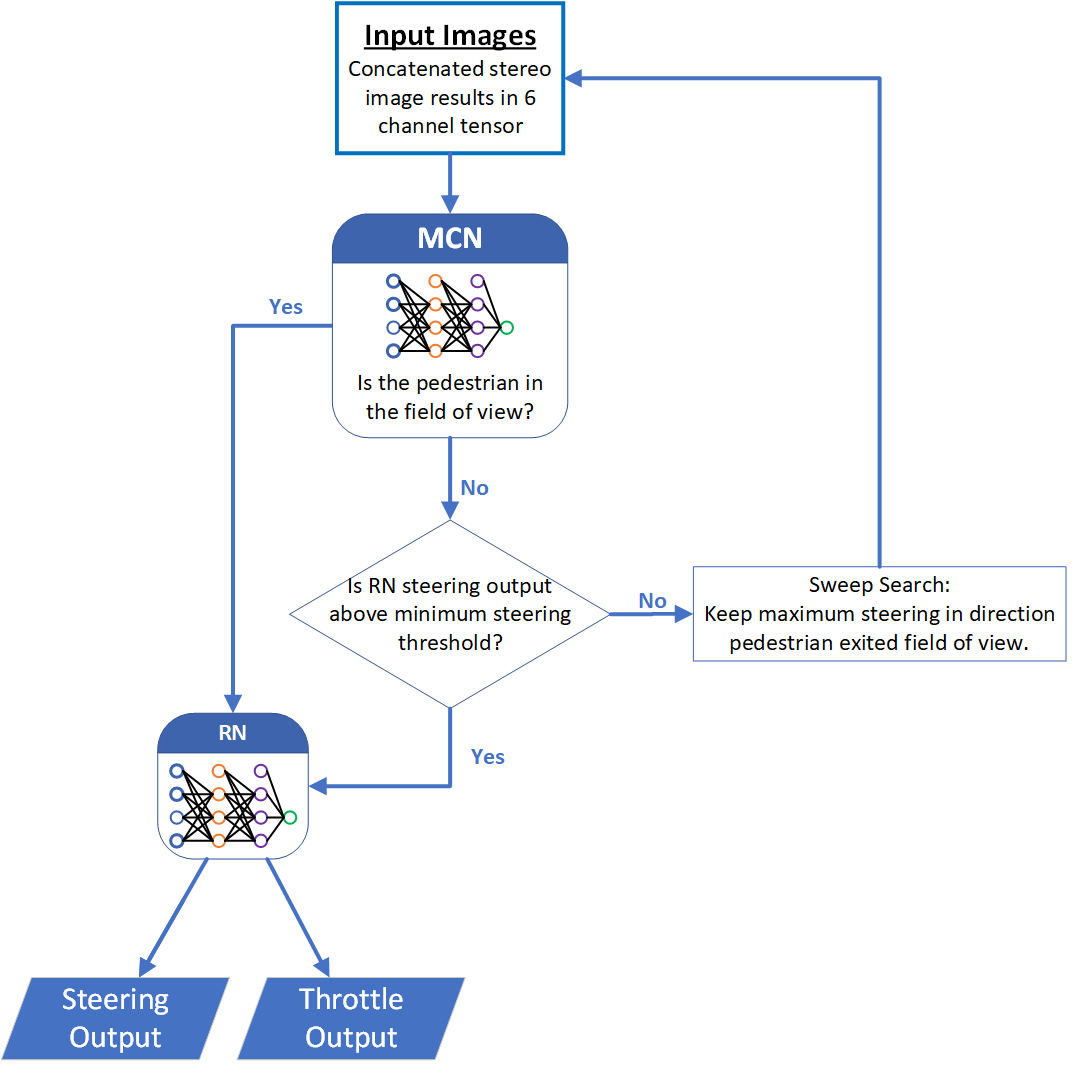}
  \caption{Hierarchical Control Framework}
  \label{fig:multiTaskUserIntent}
\end{figure}

The hierarchical framework for end-to-end control of the AV platform is documented in \figref{fig:multiTaskUserIntent}. The entry into the network is a six-channel image acquired by concatenation of the stereo-vision camera feed which is classified by the Master Classifier Network (MCN) as containing the target pedestrian or not. If it determines that the pedestrian is within the field of view, it feeds the Regression Network (RN) which in turn runs inference on the image to produce a steering angle and throttle value that will allow it to maintain cadence with the pedestrian.

If the target pedestrian is not within the field of view, then an exploratory search strategy is initiated which pivots around the inference capabilities of the RN. The RN will be described in detail shortly, but pertinent to the strategy is that an LSTM layer is incorporated close to the top of the model to give the network state memory.  If a target pedestrian walks laterally out of the field of view, the model will infer a steering value to reacquire them. If the pedestrian is not recovered within ${\sim}10$ frames of their departure, (training an LSTM for larger sequence lengths tends to create instability in its response profile), the RN will not be able to reacquire without external prompting. This external prompting will come from a control loop that monitors the steering signal: if the steering signal produced from the RN is above a minimal threshold, the MCN defers to the RN's inference output and the pedestrian is reacquired by continuous regression inference. If the RN steering value drops below a certain threshold, (indicative of the network having no certainty in which direction to turn), the control loop will use the last steering direction to force the AV into a continuous circular lap to sweep for the pedestrian. If after 10 seconds of sweep search the pedestrian is still not reacquired, then the vehicle performs an emergency stop. When the MCN determines that the pedestrian is reacquired, it feeds the RN camera video feed and regression inference once again determines the tracking path of the AV.

\subsection{Classifier Model}

\begin{figure}[htbp!]
  \centering
  \includegraphics[scale=0.5]{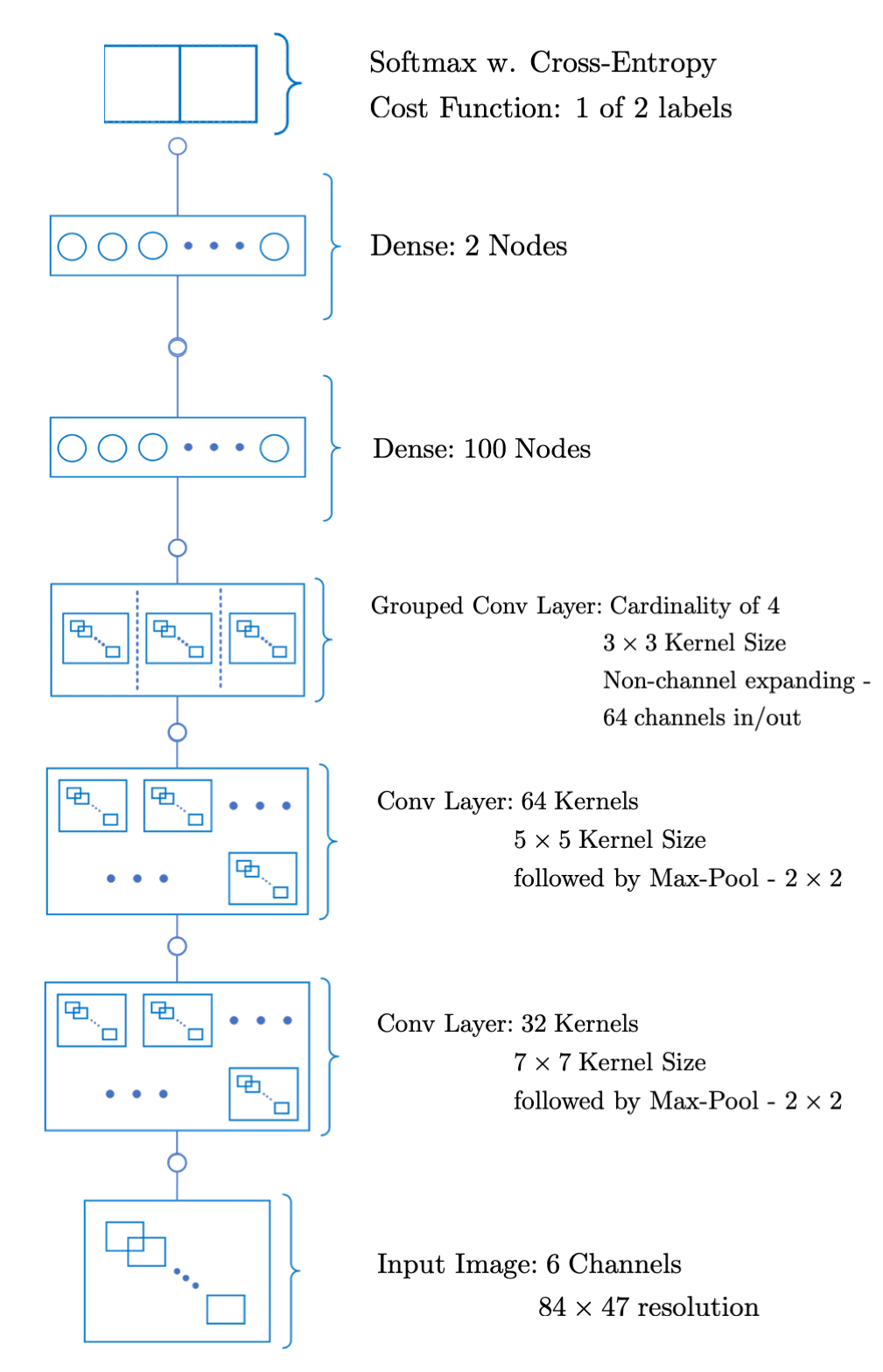}
  \caption{MCN Architecture}
  \label{fig:MCN}
\end{figure}

The MCN has a relatively straightforward but pivotal task since it serves as the entry gateway to the hierarchical framework. As such, a significant amount of experimentation went into designing the robust architecture presented in \figref{fig:MCN}.

The architecture begins with a sequence of two convolutional/max-pooling blocks, which expand the input tensor's channels from 6 to 64. After the second convolution block, a grouped convolution block is inserted with a cardinality of 4, kernel size 3 and no channel expansion.  A conceptual review of grouped convolutions is given shortly.  The grouped convolution is itself followed by a pooling layer, which feeds a sequence of two dense layers that then submit their output to a softmax layer. Binary cross-entropy could have been used as the loss function here, but experimentation with tracking other objects besides pedestrians used the same framework so conventional cross-entropy was used for training.

It is noted that batch normalization \citep{BN} was used at various stages in the network development process, but due most likely to the relatively shallow nature of the networks, no significant improvement in model performance resulted from its use despite a slight increase in compute latency, so it was not included in the final MCN or RN architectures.

\newpage

\subsection{Regression Model}

The RN consists of certain common elements with the MCN, with the entry of the network consisting of a sequence of three convolutional blocks feeding into a grouped convolution. An LSTM layer is inserted post grouped convolution to build in state memory for the network. Using 60 hidden nodes and a sequence of 5 frames, the LSTM's primary role is to correct for sudden lateral movements of the target pedestrian, allowing for quick recovery of the AV in the appropriate direction if the pedestrian were to suddenly depart from the field of view. An additive L$_2$ cost function is applied across the steering and throttle node outputs of the dense layer that sits at the top of the network.

\begin{figure}[htb]
  \centering
  \includegraphics[scale=0.5]{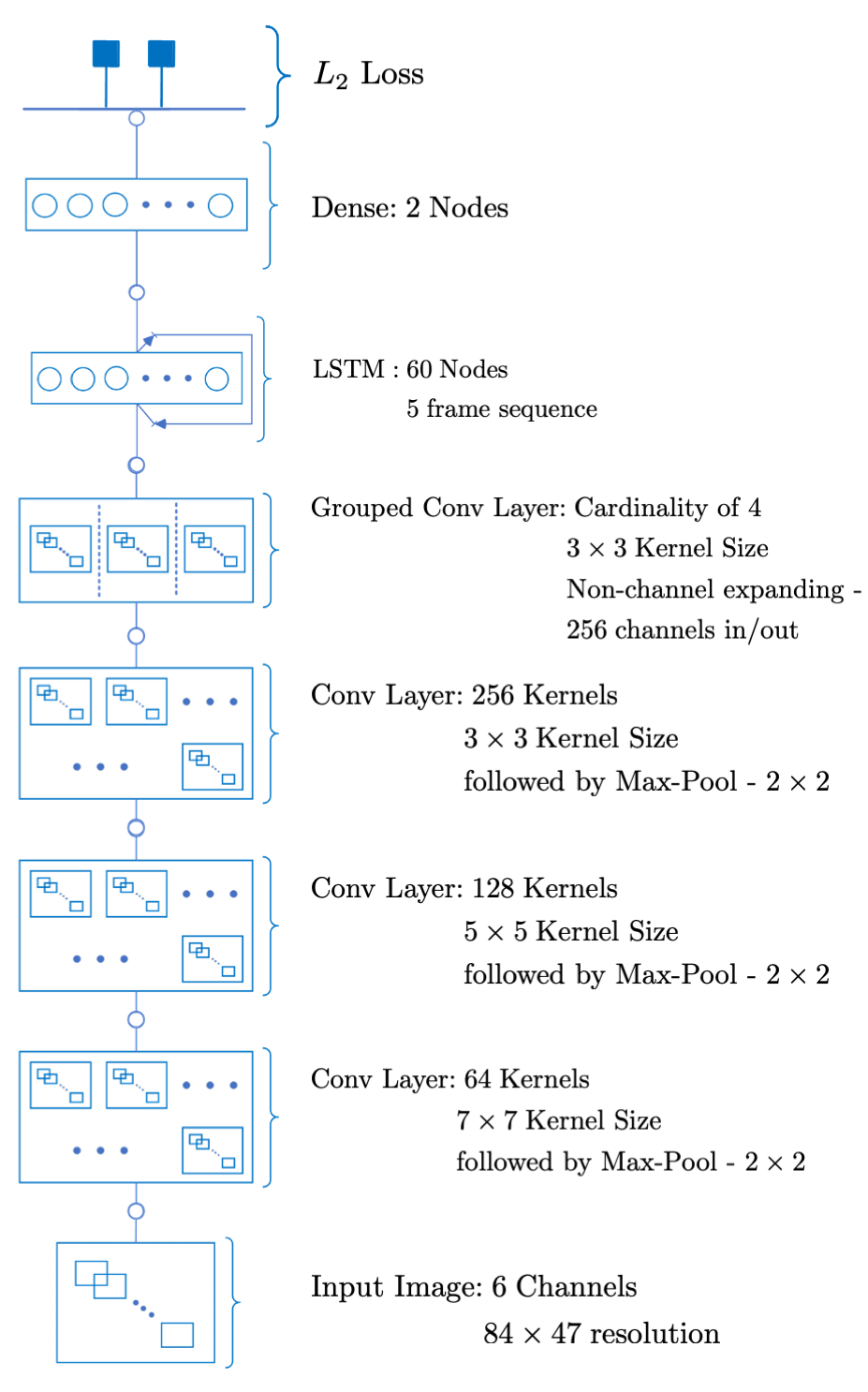}
  \caption{RN Architecture}
  \label{fig:RN}
\end{figure}

\newpage
\subsection{Grouped Convolutions}

\begin{figure}[htbp!]
  \centering
  \includegraphics[scale=0.3]{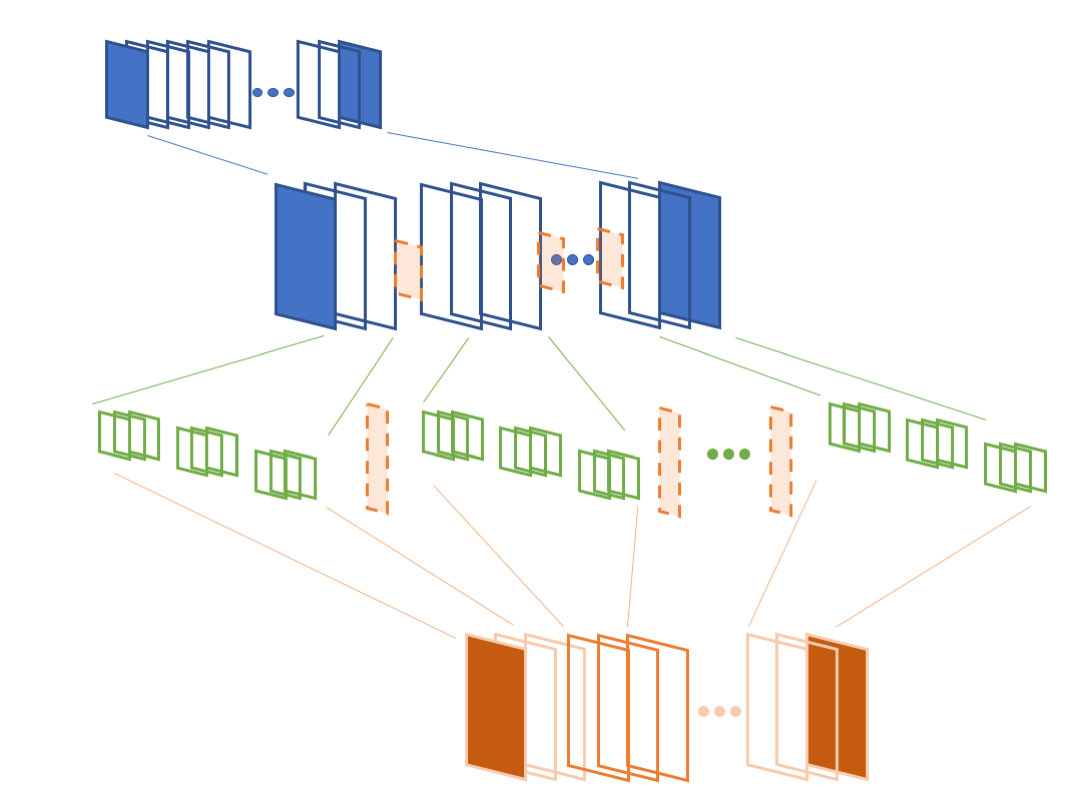}
  \caption{Grouped Convolutions}
  \label{fig:group_conv}
\end{figure}

Despite the MCN and RN performing distinct roles in the hierarchical framework, both network architectures have benefited from the introduction of grouped convolutions. As Xie et al \citep{ResNeXt} point out, one of the first most widely circulated introductions to grouped convolutions resulted out of computational necessity since the authors of \textit{AlexNet} \citep{AlexNet} were striving to train an image classifier across two Graphical Processing Units (GPUs) and in so doing wanted to share certain convolutional layers' workloads across devices. It was not until some years later that Xie et al proposed that  grouped convolutions should be revisited as a means to directly boost the representational capabilities of convolutions, and in so doing produced the highly versatile and accurate \textit{ResNeXt} model. The impact of this novel layer are even now being explored in NLP applications, where Iandola et al \citep{SB} have demonstrated the significant computational savings they provide once introduced into the transformer block model. There are a few versions of grouped convolutions introduced in the literature, but what was implemented here is the non-channel expanding variety which is conceptually rendered in \figref{fig:group_conv}.

A given input tensor of arbitrary channel width $P$ is segregated into distinct \textit{groups} consisting of $(P/n_{\text{groups}})$ channels. To each of these \textit{groups}, a separate set of convolutional kernels are applied.  These \textit{grouped convolutions} produce output feature maps that maintain the channel width dimension and feed their output feature maps to a concatenation across all groups, producing a final tensor of the same channel width as the input tensor. Let's consider the discrete, two-dimensional cross-correlation operator at the heart of convolutional layers, 

\begin{equation}
f[x,y] \; = \mathcal{I} * \mathcal{K} = \; \sum\limits_{m=-\infty}^{\infty} \sum\limits_{n=-\infty}^{\infty} \mathcal{I}[x+m, y+n] \; \mathcal{K}[m, n]
\end{equation}
where $\mathcal{I}$ and $\mathcal{K}$ are the input tensor and filter kernel respectively. If we consider this application across an input tensor of $P$ channels, a single element of the output feature map $z$ would result from 

\begin{equation}
o[x,y,\text{channel} \; z] \; = \; \sum\limits_{p=1}^{P}  \left[ \sum\limits_{m=-\infty}^{\infty} \sum\limits_{n=-\infty}^{\infty} \mathcal{I} [x+m, y+n] \; \mathcal{K} [m, n]\right]_p
\end{equation}

If now the input channel dimension is divided into $G$ groups, an element of the output feature map becomes

\begin{equation}
o[x, y,\text{channel} \; z] \; = \; \sum\limits_{p=1}^{\frac{P}{G}}  \left[ \sum\limits_{m=-\infty}^{\infty} \sum\limits_{n=-\infty}^{\infty} \mathcal{I}[x+m, y+n] \; \mathcal{K}[m, n]\right]_p
\end{equation}

Finally, if one concatenates the output feature maps across all groups, (as done in \figref{fig:group_conv}), 

\begin{equation}
O \; = \; O_{\text{group 1}} \; ^\frown \;  O_{\text{group 2}} \; ^\frown \; \dots ^\frown \; O_{\text{group G}}
\end{equation}

then in effect one is reducing the compute required by the predefined number of groups. As Iandola et al \citep{SB} adeptly point out, the ultimate effect is to reduce the flop count per convolution by $G$.

\tableref{table:comparing_gc} summarizes  the impact of introducing grouped convolutions into the MCN and RN architectures by replacing the grouped convolution layers with conventional convolutions and then comparing the trainable variable counts and the floating point operations.

\begin{table}[htbp!]
  \caption{Comparing Grouped Convolutions with Conventional Convolutions}
  \label{table:comparing_gc}
  \centering
  \begin{tabular}{llll}
    \toprule
      RN & w Standard Convs & w Grouped Convs & \% Reduction\\
    \midrule
	Trainable Parameters        	& 2.23M			& 1.79M    & 19.84    \\
	Floating Point Operations* 	& 3.95B   			&	3.66B    &  7.36        \\
	\midrule \\
	\midrule 
	   MCN   & w Standard Convs & w Grouped Convs &  \% Reduction\\
    \midrule
	Trainable Parameters        &	 5.20K  		 &   4.93K     & 5.31    \\
	Floating Point Operations  &  200M           &   186M      &  7.06       \\
    \bottomrule \\
    \multicolumn{4}{c}{* Networks are run in inference state using Tensorflow \citep{tensorflow} metric tools} 
  \end{tabular} \\
\end{table}

The open question is why does segregating the channels of the input tensor actually improve representation? This is much more difficult to ascertain. Xie et al document their experiments with \textit{ResNeXt}, correlating greater classification performance of their model with cardinality, i.e. number of distinct groups. A key facet that might impact performance is that in conventional convolutional layers, each resulting feature map results from aggregating the filter kernels' output across all input channels. By segregating the input channels when applying grouped convolutions, each individual channel is more heavily weighted in the resulting output tensor. Further investigation is warranted into identifying the underlying benefits.

\subsection{Comparing Training Performance to Graphical Processing Units}

Prior to this phase of work, the models ported to the scaled AV were exclusively trained on GPUs \citep{MULTI}, and so a basic performance comparison in terms of training was made with the final versions of the networks on both the IPU and GPU.  It is noted that the results presented in \tableref{table:training_comparison} were done using the same batch-sizes and hyper-parameter configurations to produce a direct  comparison. At the time this work was conducted, the team involved at Ford Research used NVIDIA's Titan X cards with the Maxwell chip and therefore that platform was used in comparing performance throughput to that of the IPU Mark 1. Newer architectures for both the GPU and IPU platforms are now currently available and future work will be done to compare performance metrics on the upgraded chip sets. Larger batch sizes would have boosted GPU performance, but the objective was to compare based on how the models were actually trained for deployment given that larger batch sizes tended to have less stable loss profiles for the given optimizer selected, which in this case was SGD. The results presented here were averaged over 10 epochs of training on actual data. To fully utilize the IPU's C2 card \citep{CITADEL}, models were run with a replication factor of two; i.e. the model was deployed to both IPUs of the same card and run in a data parallel fashion.

\begin{table}[htbp!]
  \caption{Training Throughput Comparison}
  \label{table:training_comparison}
  \centering
  \begin{tabular}{lllll}
    \toprule
     Model & Batch Size & Throughput on IPU (images/sec)  &  on GPU  & $\times$ Speed-up \\
    \midrule
	MCN       	& 8	& 5148  & 1360 & 3.78   \\
	RN     		& 1   & 2914  &  413  & 7.05   \\
    \bottomrule \\
  \end{tabular} \\
\end{table}

\section{Creating a Training Set}

The scaled AV platform facilitates the process of creating training data for a variety of tasks. When remotely controlled by a human operator,  the AV logs video feed from the stereo camera in addition to input from the operator in terms of steering and throttle. These controls are logged as PWM signals, which an RC transmitter sends to the micro-controller that manages the steering servo and the ESC of the motor.  It is noted that both steering and throttle signals are normalized and scaled to have a range of $-100$ to $100$ in the case of steering and $0$ to $100$ in the case of throttle.  For each frame, the steering commands and throttle adjustments made by the driver are logged concurrently, so that a network can then be trained based on the image data to replicate the steering and throttle responses of the human driver. The HDF5 \citep{HDF5} format is to used to keep all distinct frame data sequentially logged.

Building on this capability, the strategy to train the hierarchical framework is as follows. A pedestrian stands at four distinct locations of a circle: imagining a clock face, the pedestrian starts from 12 o'clock, 3 o'clock, 6 o'clock and 9 o'clock. The scaled AV platform records a fixed set of frames of the pedestrian standing in front of the camera at each of these locations, and then a fixed number of frames of no pedestrian in the camera's field of view. This data is appropriately labeled and used to train the classifier network. A subset set of frames, ${\sim}20\%$, are set aside as a test set. The pedestrian then once again steps within the field of view of the camera and starting from the four distinct locations completes a circular lap while a driver remotely controls the AV. Four laps are recorded in the clockwise and counter-clockwise direction, with two additional laps set aside as a test set. 

Two distinct pedestrians were used to create the data, so a total of 8 laps in each direction (16 laps total) are used for final training. The training data for both pedestrians is mixed for the MCN as well as the RN, and performance is quantified based solely on performance of inference on the test data set.

\subsection{The Scaled AV Platform}

\begin{figure}[htp]
  \centering
 \includegraphics[scale=0.28]{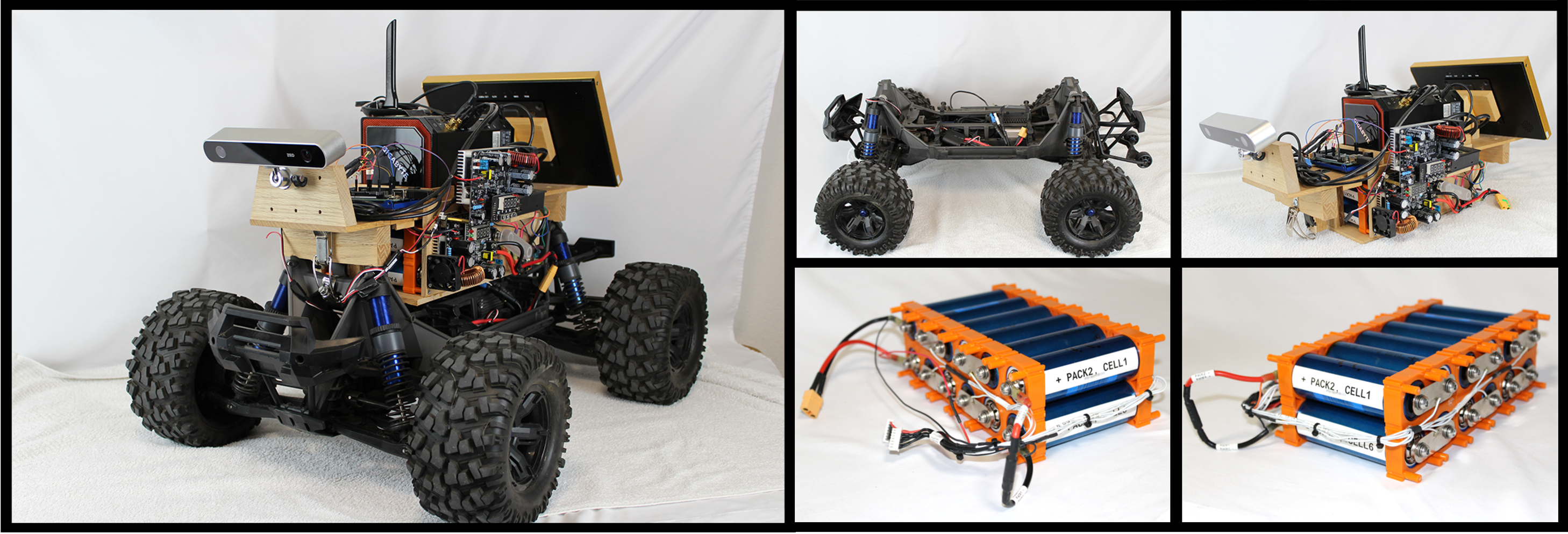}
  \caption{Scaled-vehicle platform consisting of 1/6$^{th}$ scaled platform with Zed stereo-camera, Arduino Mega 2560 micro-controller, NUC running Ubuntu 16.04, 12 cell LiFePO4 battery pack and wireless transmitter to record steering input from driver.}
  \label{fig:xMaxx}
\end{figure}

The scaled AV platform used for the work is described in detail in \citep{MULTI}.  One notable improvement since that work is an upgrade to the battery pack from lead-acid to a 12 cell "M" size LiFePO4 battery, which significantly increased the operational time of the AV to about 3 to 4 hours. The new battery pack is shown in \figref{fig:xMaxx}.

Of specific note here is that the scaled AV's camera is a Stereolab's\textsuperscript{\tiny{\textregistered}} stereo-vision camera, and so camera sensor feed consists of two RGB image frames which are concatenated to create a six-channel input image for each network.

\section{Training}

The MCN and RN architectures were defined and trained in Tensorflow/Keras using the IPU. The MCN required binary labeling to recognize when a pedestrian was present in the field of view, while the RN required steering and throttle values with each frame to learn the appropriate vehicle response.  The MCN used shuffled image frames from 16 log files, each labeled in terms of depicting a pedestrian or not, and stochastic gradient with a static learning rate was applied to a cross-entopy loss. The RN required a sequence of 5 image frames to train the LSTM, and used shuffled data sequences from 18 distinct log files. An additive $L_2$ loss was used across the RN's steering and throttle output nodes.  A summary of the training schedule is given in \tableref{table:training_schedule}.

\begin{table}[htbp!]
  \caption{Training Schedule for the MCN and RN}
  \label{table:training_schedule}
  \centering
  \begin{tabular}{lllll}
    \toprule
    Model & Base Log Files & Training Data Frames & Loss/Optimizer & Epochs \\
    \midrule
	RN        	   & 18  						&   31,440                 &  Additive $L_2$ w. SGD          & 500 \\
	MCN  		   & 16						    &   13,639                 &   Cross-entroy w. SGD            & 100 \\
    \bottomrule \\
  \end{tabular} \\
\end{table}

\section{Results}

Results are presented in two distinct sets of quantification: the accuracy of the MCN as a classifier and the accuracy of the RN as a regression model for steering and throttle given that these are the primary indicators of how the hierarchical framework will operate overall. 

\subsection{MCN Classifier Performance}

\begin{figure}[htp]
  \centering
 \includegraphics[scale=0.3]{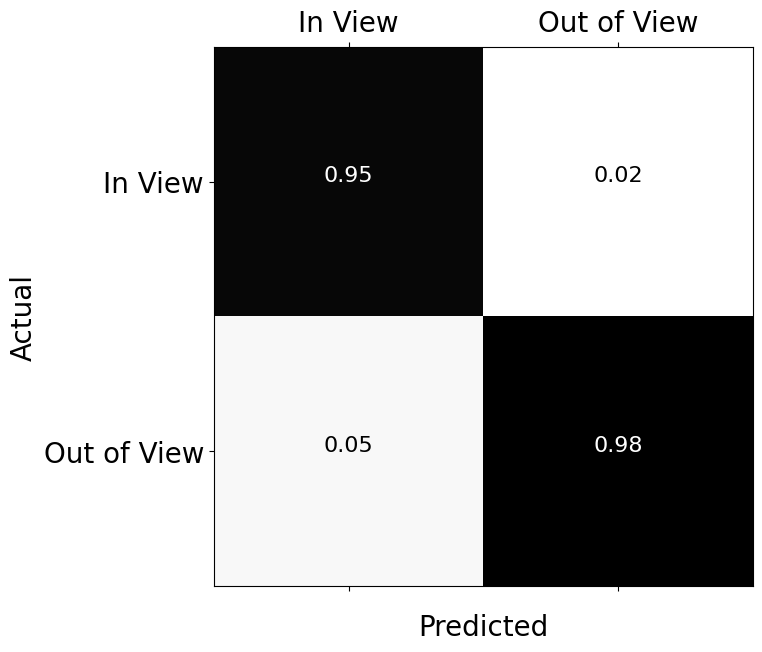}
  \caption{MCN confusion matrix normalized based on precision}
  \label{fig:conf_matrix}
\end{figure}

The MCN can be trained to a very high degree of accuracy and the confusion matrix resulting on the test data set is presented in \figref{fig:conf_matrix}. The inference classification were normalized for precision.

\subsection{RN Regression Performance}

\begin{figure} [htp]
\centering
\begin{tabular}{cc}
\includegraphics[width=0.5\textwidth]{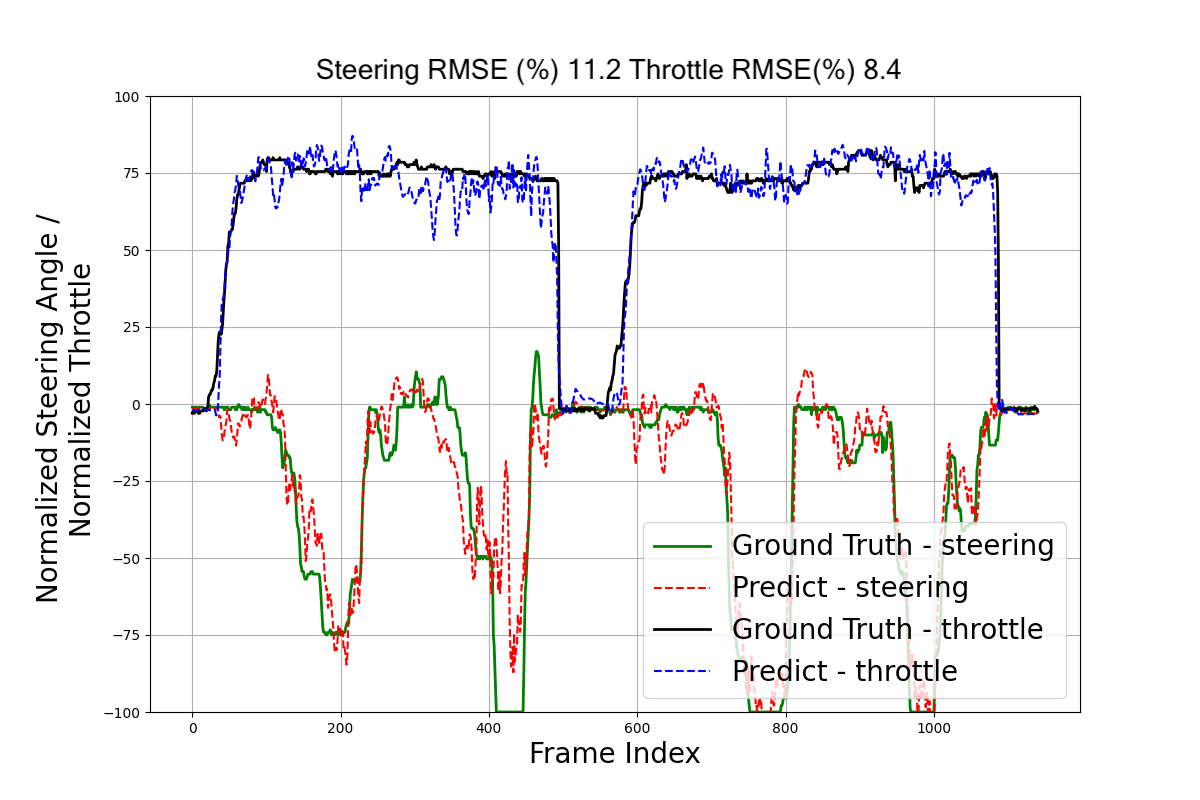} &
\includegraphics[width=0.5\textwidth]{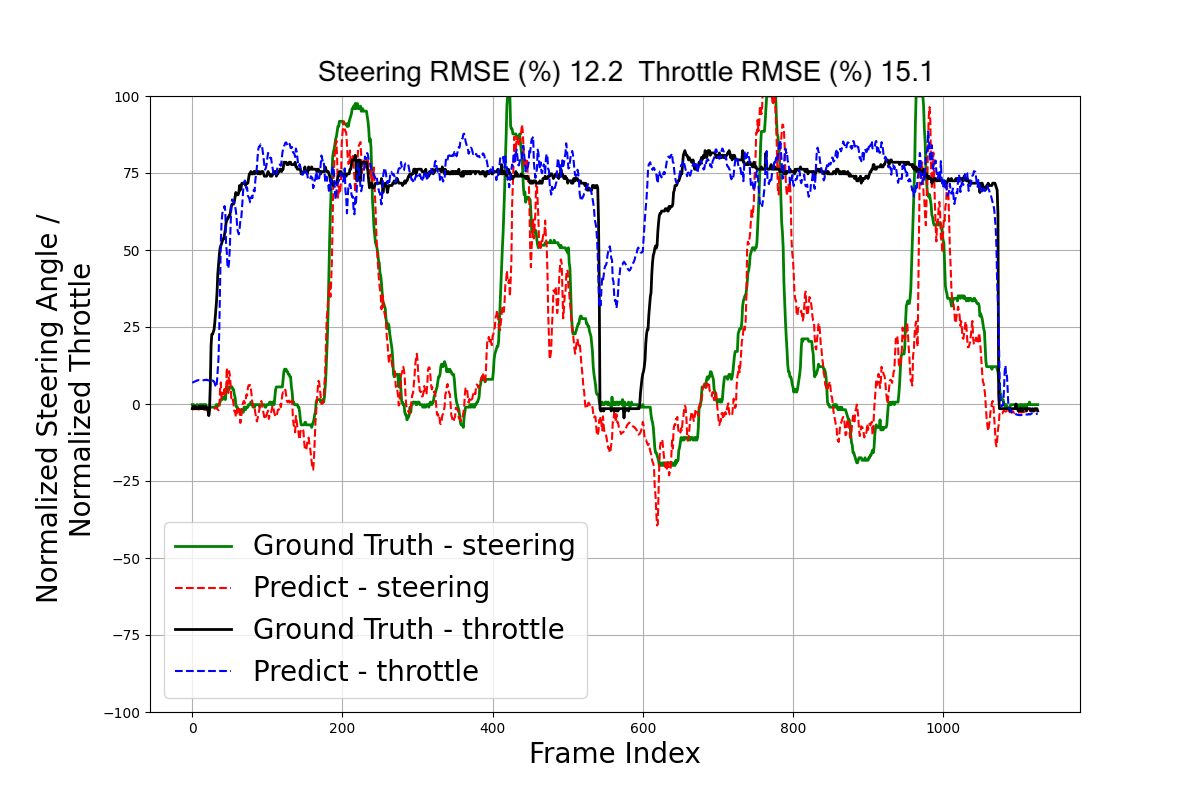} \\
{(a) Pedestrian A: counter-clockwise turn}  & {(b) Pedestrian A: clockwise turn}  \\[4pt]
\end{tabular}
\begin{tabular}{cc}
\includegraphics[width=0.5\textwidth]{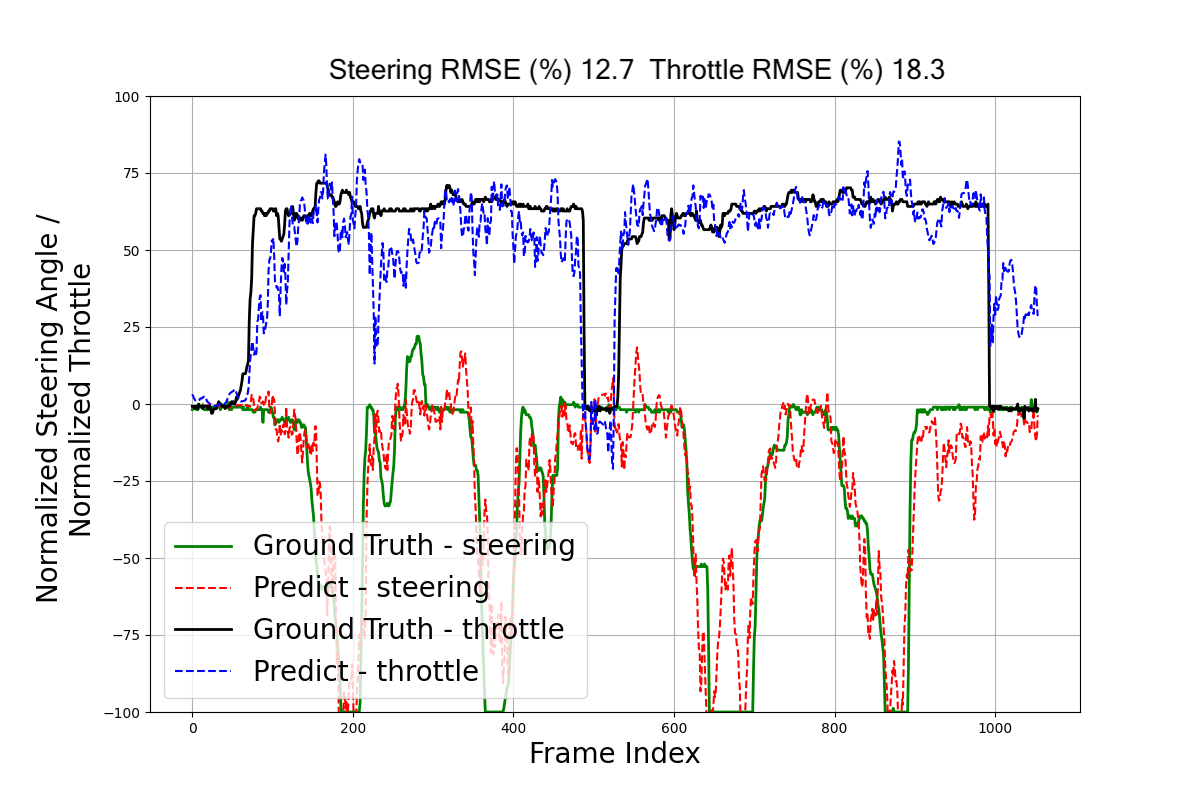} &
\includegraphics[width=0.5\textwidth]{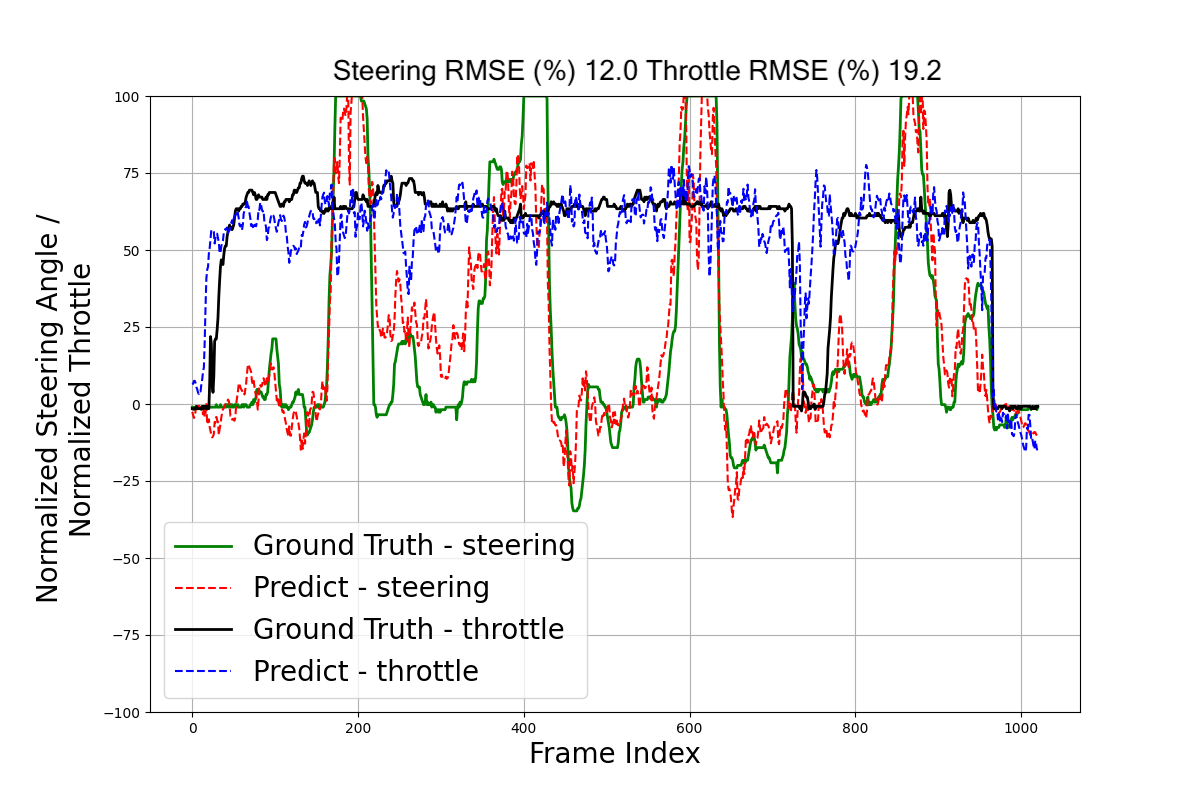} \\
{(c) Pedestrian B: counter-clockwise turn}  & {(d) Pedestrian B: clockwise turn}  \\[4pt]
\end{tabular}
\caption{Inference performance on test data of RN for the two distinct pedestrians. }
\label{fig:RN_performance}
\end{figure}

The performance of the RN are quantified by two test laps in each direction for both pedestrian subjects and are presented in \figref{fig:RN_performance}. It is noted that although the trajectory of the test laps are similar to the training set, there is significant variation between each lap that is logged since it relies on manual input from the driver in terms of throttle and steering; i.e. the ground truth has some variability. Thus the resulting steering and throttle root-mean-square error (RMSE) values are not a completely concrete metric, but based on extensive experimentation,  RMSE on the order of $20\%$ and below indicate successful regression performance.  The RMSE reported here is given as a percentage of the total normalized throttle and steering ranges.  It is noted that a moving average of 10 frames was used to modulate the throttle output during real-time inference in live drive tests to minimize jolting behavior of the scaled-vehicle due to sudden throttle value changes. The moving average is computed as

\begin{equation}
  throttle_{t+1} = \frac{\sum\limits_{n=t-N}^{t} throttle_{n}} {N}
\end{equation}

where $N = 10$ and the throttle value at time step $t + 1$ is the average value of the previous $t - N$ time steps. The delay associated with this moving average of  10 frames had no visible impact on the overall performance of the AV.

\newpage

\subsection{A Random Walk}

As a final drive test, one of the pedestrians was requested to walk in a random trajectory along a path that was not used for training or testing and the resulting behavior was recorded and posted here:  \href{https://www.youtube.com/watch?v=1IBgvzCI0LQ&feature=youtu.be}{Random Trajectory Walk Video} \footnote{Video link: https://www.youtube.com/watch?v=1IBgvzCI0LQ\&feature=youtu.be}. Due to wildfire conditions in the vicinity of where this drive test was conducted, an additional data set for training the classifier was required to account for the uncharacteristically smokey and hazy lighting conditions. 

\section{Conclusion}

The work has presented a flexible framework to train autonomous systems to leverage the representation capabilities of DNNs to perform the task of following-the-leader. The hierarchical paradigm used in the authors' previous work lent itself directly to the current objective, and by using the IPU as the training platform, the authors were able to rapidly prototype and experiment with numerous network architectures achieving a significant boost to training throughput, (on the order of a factor of 3.5 for the MCN and 7 for the RN), while producing very robust steering and throttle profiles from the control framework.  In terms of future work, the authors aim to replicate this effort on actual factory floors involved in the production of vehicles for the Ford Motor Company so as to enhance the capability and applicability of the control framework presented. 

\section{Acknowledgments}

The authors would like to thank Venkatapathi Nallapa of Greenfield Labs, (Ford Motor Company), for his valuable support and encouragement during the effort to conduct this research.

\section{Conflict of Interest}

On behalf of all authors, the corresponding author states that there is no conflict of interest.

\bibliographystyle{unsrtnat}
\bibliography{references}

\begin{thebibliography}{12}
\providecommand{\natexlab}[1]{#1}
\providecommand{\url}[1]{\texttt{#1}}
\expandafter\ifx\csname urlstyle\endcsname\relax
  \providecommand{\doi}[1]{doi: #1}\else
  \providecommand{\doi}{doi: \begingroup \urlstyle{rm}\Url}\fi

\bibitem[Goodchild and Toy(2017)]{Delivery}
Anne Goodchild and Jordan Toy.
\newblock Delivery by drone: An evaluation of unmanned aerial vehicle
  technology in reducing co 2 emissions in the delivery service industry.
\newblock \emph{Transportation Research Part D: Transport and Environment}, 61,
  03 2017.
\newblock \doi{10.1016/j.trd.2017.02.017}.

\bibitem[Relles and Patterson(2011)]{AUVs}
Noelle~J. Relles and Mark~R. Patterson.
\newblock \emph{AUVs (ROVs)}, pages 71--75.
\newblock Springer Netherlands, Dordrecht, 2011.
\newblock ISBN 978-90-481-2639-2.
\newblock \doi{10.1007/978-90-481-2639-2_40}.
\newblock URL \url{https://doi.org/10.1007/978-90-481-2639-2_40}.

\bibitem[Klančar et~al.(2009)Klančar, Matko, and Blazic]{laser}
Gregor Klančar, Drago Matko, and Saso Blazic.
\newblock Wheeled mobile robots control in a linear platoon.
\newblock \emph{Journal of Intelligent and Robotic Systems}, 54:\penalty0
  709--731, 05 2009.
\newblock \doi{10.1007/s10846-008-9285-7}.

\bibitem[{Baarath, K.} et~al.(2017){Baarath, K.}, {Zakaria, Muhammad Aizzat},
  {Suparmaniam, M.V.}, and {Abu, Mohd Yazid bin}]{Kinect}
{Baarath, K.}, {Zakaria, Muhammad Aizzat}, {Suparmaniam, M.V.}, and {Abu, Mohd
  Yazid bin}.
\newblock Platooning strategy of mobile robot: simulation and experiment.
\newblock \emph{MATEC Web Conf.}, 90:\penalty0 01060, 2017.
\newblock \doi{10.1051/matecconf/20179001060}.
\newblock URL \url{https://doi.org/10.1051/matecconf/20179001060}.

\bibitem[Solomon and cois Charette(2019)]{MULTI}
Jos\'e Solomon and Fran\c cois Charette.
\newblock Hierarchical multi-task deep neural network architecture for
  end-to-end driving.
\newblock \emph{arXiv: CoRR}, abs/1902.03466, 2019.
\newblock URL \url{https://arxiv.org/abs/1902.03466}.

\bibitem[Krizhevsky et~al.(2012)Krizhevsky, Sutskever, and Hinton]{AlexNet}
Alex Krizhevsky, Ilya Sutskever, and Geoffrey~E Hinton.
\newblock Imagenet classification with deep convolutional neural networks.
\newblock In F.~Pereira, C.~J.~C. Burges, L.~Bottou, and K.~Q. Weinberger,
  editors, \emph{Advances in Neural Information Processing Systems 25}, pages
  1097--1105. Curran Associates, Inc., 2012.
\newblock URL
  \url{http://papers.nips.cc/paper/4824-imagenet-classification-with-deep-convolutional-neural-networks.pdf}.

\bibitem[Xie et~al.(2017)Xie, Girshick, Dollar, Tu, and He]{ResNeXt}
Saining Xie, Ross Girshick, Piotr Dollar, Z.~Tu, and Kaiming He.
\newblock Aggregated residual transformations for deep neural networks.
\newblock \emph{10.1109/CVPR.2017.634}, pages 5987--5995, 07 2017.

\bibitem[Jia et~al.(2019)Jia, Tillman, Maggioni, and Scarpazza]{CITADEL}
Zhe Jia, Blake Tillman, Marco Maggioni, and Daniele~Paolo Scarpazza.
\newblock Dissecting the graphcore ipu architecture via microbenchmarking.
\newblock \emph{arXiv}, 1912.03413, 2019.

\bibitem[Ioffe and Szegedy(2015)]{BN}
Sergey Ioffe and Christian Szegedy.
\newblock Batch normalization: Accelerating deep network training by reducing
  internal covariate shift.
\newblock \emph{arXiv}, 2015.

\bibitem[Iandola et~al.(2020)Iandola, Shaw, Krishna, and Keutzer]{SB}
Forrest~N. Iandola, Albert~E. Shaw, Ravi Krishna, and Kurt~W. Keutzer.
\newblock Squeezebert: What can computer vision teach nlp about efficient
  neural networks?
\newblock \emph{CoRR}, abs/2006.11316, 2020.
\newblock URL \url{https://arxiv.org/abs/2006.11316}.

\bibitem[ten(2015)]{tensorflow}
{TensorFlow}: Large-scale machine learning on heterogeneous systems, 2015.
\newblock URL \url{https://www.tensorflow.org/}.
\newblock Software available from tensorflow.org.

\bibitem[Group(2010)]{HDF5}
The~HDF Group.
\newblock Hierarchical data format version 5, 2010.
\newblock URL \url{http://www.hdfgroup.org/HDF5}.

\end{thebibliography}

\end{document}